\documentclass[lettersize,journal]{IEEEtran}
\usepackage{amsmath,amsfonts}
\usepackage{changes}
\usepackage{algorithm}  

\usepackage{algorithmicx}
\usepackage{algpseudocode}

\usepackage{array}
\usepackage[caption=false,font=normalsize,labelfont=sf,textfont=sf]{subfig}
\usepackage{textcomp}
\usepackage{stfloats}
\usepackage{url}
\usepackage{verbatim}
\usepackage{graphicx}
\usepackage{cite}
\usepackage{xcolor}
\usepackage{array, makecell}
\usepackage{amssymb}
\begin{document}

\title{Exploring Winograd Convolution for Cost-effective Neural Network Fault Tolerance}

\author{Xinghua Xue,
        Cheng Liu,
        Bo Liu, 
        Haitong Huang,
        Ying Wang,
        Tao Luo, \\
        Lei Zhang,
        Huawei Li,~\IEEEmembership{Senior Member,~IEEE}
        Xiaowei Li,~\IEEEmembership{Senior Member,~IEEE}

\thanks{Corresponding author: Cheng Liu.}
\thanks{This article was presented in part
at the 59th IEEE International Conference on Design Automation Conference (DAC), July 10–24, 2022.}
\thanks{Xinghua Xue, Cheng Liu, Haitong Huang, Ying Wang, Lei Zhang, Huawei Li, and Xiaowei Li are with both State Key Lab of Processors, Institute of Computing Technology, Chinese Academy of Sciences, Beijing 100190, China, and University of Chinese Academy of Sciences, Beijing 100190, China. Bo Liu is with Beijing Institute of Control Engineering. Tao Luo is with Institute of High Performance Computing, A*STAR, Singapore. (e-mail: \{xuexinghua, liucheng\}@ict.ac.cn).}
}

\markboth{IEEE TRANSACTIONS ON VERY LARGE SCALE INTEGRA TION (VLSI) SYSTEMS, VOL. XX, NO. XX, XXX 2023}%
{Xinghua Xue \MakeLowercase{\textit{et al.}}: Bare Demo of IEEEtran.cls for IEEE Journals}

\maketitle

\begin{abstract}
Winograd is generally utilized to optimize convolution performance and computational efficiency because of the reduced multiplication operations, but the reliability issues brought by winograd are usually overlooked. In this work, we observe the great potential of winograd convolution in improving neural network (NN) fault tolerance. Based on the observation, we evaluate winograd convolution fault tolerance comprehensively from different granularities ranging from models, layers, and operation types for the first time. Then, we explore the use of inherent fault tolerance of winograd convolution for cost-effective NN protection against soft errors. Specifically, we mainly investigate how winograd convolution can be effectively incorporated with classical fault-tolerant design approaches including triple modular redundancy (TMR), fault-aware retraining, and constrained activation functions. According to our experiments, winograd convolution can reduce the fault-tolerant design overhead by 55.77\% on average without any accuracy loss compared to standard convolution, and further reduce the computing overhead by 17.24\% when the inherent fault tolerance of winograd convolution is considered. When it is applied on fault-tolerant neural networks enhanced with fault-aware retraining and constrained activation functions, the resulting model accuracy generally shows significant improvement in presence of various faults.
\end{abstract}


\section{Introduction}

\IEEEPARstart{I}{nspired} by the great success of deep neural networks (DNNs) in computer vision and natural language processing in the past decade \cite{hussain2022design}, people seek to explore the use of DNNs in more and more domains of applications. Many of the applications such as autonomous driving and medical robotics can be closely coupled with human safety and demand resilient processing of DNNs. Otherwise, unexpected inference failure may even lead to catastrophic consequences \cite{yampolskiy2016artificial} \cite{hanif2020dependable}. While the underlying computing engines that sustain DNN processing are typically fabricated with nanoscale technologies recently, and can inevitably suffer growing soft errors induced by alpha particles, neutrons, and heavy ions \cite{bohr2017cmos, Pandey2019GreenTPU, goodman1991reliability, zhu2004effects, mastipuram2004soft}, fault-tolerant design approaches against soft errors become critical to the adoption of DNNs on these safety-critical applications. In addition, many prior works \cite{reagen2018ares, torres2017fault, temam2012defect, reagen2016minerva} demonstrate that the fault tolerance of DNNs can also be utilized to relax the requirements of 100\% correctness of the execution, and leveraged for higher performance and energy efficiency through techniques such as voltage scaling \cite{safari2022survey} \cite{zhang2018thundervolt}, overclocking \cite{chen2023improving} \cite{jiang2023automated} and model pruning \cite{deng2020model}.

\IEEEpubidadjcol
In order to improve the fault tolerance of DNNs, various approaches have been proposed from different perspectives. Classical triple modular redundancy (TMR) \cite{lyons1962use} is a straightforward way to enhance the fault tolerance of DNNs yet will induce more than 200\% computing overhead. Selective TMR or hardening \cite{libano2018selective} \cite{schorn2018accurate} is generally preferred, as it significantly reduces the redundancy overhead by protecting only a fraction of the most vulnerable part of the DNN processing. Prior work in \cite{kosaian2021arithmetic, xue2023approxabft, zhao2020ft} applied typical algorithm-based fault tolerance (ABFT) algorithm to actively perform a light-weight error detection approach via a matrix-matrix multiplication based checksum mechanism. On top of the error detection, error correction which is usually much more expensive will only be invoked when errors are detected. In this case, the error correction overhead can be greatly amortized across the entire neural network processing. The above approaches improve the DNN fault tolerance significantly, but they generally require additional error detection overhead at runtime and usually affect the DNN data flow and performance accordingly. Unlike the redundancy and recomputing based fault-tolerant design approaches, another categories of approaches attempt to investigate the inherent fault tolerance of DNNs. Fault-aware retraining approach \cite{huang2022bit} \cite{zahid2020fat} essentially learns fault information to adapt to the specific failure scenarios. However, retraining is required for different failure scenarios and can induce considerable training cost. Instead of tuning the model parameters, the authors in \cite{li2020ftt} \cite{ning2021ftt} utilize network architecture search (NAS) to search for fault-tolerant, high-precision, and high-performance networks. Compared to clean neural networks, the searched fault-tolerant models typically will be more complex due to the fault-tolerant design constraints, resulting in additional computing overhead or lower accuracy.

Different from prior research works, we aim to improve the inherent fault tolerance of DNNs without compromising the model accuracy nor affecting the execution performance. We observe that bit-flip errors in multiplication and addition operations usually induce distinct influence on output results. Basically, compared to addition, bit-flip errors on multiplication will generally cause larger computing bias, which makes multiplication more sensitive to soft errors accordingly. At the same time, we notice that winograd convolution can \cite{lavin2016fast} greatly reduces the number of multiplication operations in standard convolution through equivalent linear transformation \cite{lu2017evaluating, liu2018efficient, shen2018towards, jia2018optimizing}, which poses great potential in improving the fault tolerance of DNNs. With the above observations, we investigate the fault tolerance of winograd-based DNNs comprehensively, and explore how we can leverage the inherent fault tolerance for more cost-effective protection against soft errors.

Fault injection is widely utilized for fault tolerance analysis and existing fault injection tools usually inject random bit-flip errors with the granularity of neurons \cite{reagen2018ares} \cite{mahmoud2020pytorchfi} \cite{chen2020tensorfi}. While winograd convolution can be viewed as an equivalent computing approach of standard convolution with the proposed kernel partition \cite{huang2020dwm}, the output neurons of winograd convolution will be the same with that calculated with standard convolution. In this case, we cannot differentiate the fault tolerance of winograd convolution and standard convolution given the widely utilized neuron-level fault injection tools, which hinders the fault tolerance analysis of winograd convolution consequently. To address the problem, we seek to conduct more fine-grained fault injection and propose an operation-level fault injection approach, which can be aware of the different convolution calculation approaches without compromising the generality of the fault injection.

With the operation-level fault injection, we investigate the fault tolerance of winograd-based DNNs from different granularities such as models, layers, and operation types. The investigation confirms the much higher resilience of winograd convolution over standard convolution. Then, we leverage winograd convolution for more cost-effective fault tolerance of DNNs. Specifically, we explore the combination of winograd convolution with classical fault-tolerant neural network computing approaches including TMR, fault-aware retraining, and constrained activation functions. Our experiments demonstrate that winograd convolution can be utilized to reduce the fault-tolerant design overhead significantly and improve the model accuracy in presence of various soft errors.

The contributions of this work can be summarized as follows. The second and the third contributions are mainly extended on top of the conference paper\cite{xue2022winograd}.

\begin{itemize}

\item We discover and investigate the fault tolerance of winograd DNNs comprehensively for the first time from different granularities such as models, layers, and operation types. 

\item With the observation that the widely used neuron-level fault injection frameworks fail to differentiate the influence of soft errors on standard convolution and winograd convolution, we propose an operation-level fault injection framework for more fine-grained reliability analysis of neural network processing and have it open sourced on github\footnote{ \url{https://github.com/xuexinghua/Operation-level-FI.git}}.

\item We leverage the inherent fault tolerance of winograd convolution to optimize the classical fault-tolerant approaches including TMR, fault-aware retraining, and constrained activation functions. Our experiments further confirm the advantages brought by winograd convolution in terms of both fault-tolerant design overhead and model accuracy improvement under various fault configurations.

\end{itemize}

\section{Background and Related Work}

\subsection{Winograd Convolution}

Winograd convolution converts the matrix multiplication in standard convolution into element-wise multiplication, which is achieved by linearly transforming the input feature map and convolution kernels to a different domain of data representation. The element-wise multiplication results can be restored to the standard feature map domain with the corresponding inverse linear transformation. With the transform-calculation-inverse transformation process, the number of multiplication operations can be reduced considerably and the multiplication operations are replaced with more cost-effective addition operations, which significantly enhances the computing efficiency accordingly.

\vspace{-1mm}
\begin{equation}
Y_{k,b} =  A^{T}\left(\sum_{c\;=1}^C\left(Gg_{k,c}G^{T} \right) \odot\left(B^{T} d_{c,b}B\right)\right)A\label{eq:winograd-eq}
\end{equation}

A typical tiled winograd convolution is presented in Equation \ref{eq:winograd-eq}, where $\odot$ denotes the element-wise multiplication, $Y$ denotes the output feature, $g$ denotes the filter, $d$ denotes the input feature. The subscript $c$ and $k$ denote the $c$-th input channel and $k$-th output channel respectively, and $b$ denotes the $b$-th tile. $A$, $G$, and $B$ denote the transformation matrices of output, filter, and input respectively. For a $3 \times 3$ winograd convolution $F(2 \times 2,3 \times 3)$ operating on a $4 \times 4$ input tile generate a $2 \times 2$ output,  the number of multiplication is reduced $2.25\times$ from $2 \times 2 \times 3 \times 3 = 36$ to $(m + r - 1)^2 = (2 + 3 - 1)^2 = 16$. In general, the number of multiplication is reduced by ${(m\times r)^2/(m+r-1)}^2$. The data transform requires $32$ additions, and the inverse transform requires $24$ additions. When winograd convolution is conducted on $3 \times 3$ filter with unit stride, there will be no accuracy penalty. Even when the convolution filter and stride are larger, they can also be split to small ones according to the decomposable winograd method proposed in \cite{huang2020dwm} and ensures lossless conversion. 

Winograd convolution has been explored from various angles such as quantization \cite{li2020lance} \cite{chikin2022channel},   tiling \cite{shen2019toward}, hardware acceleration \cite{lu2018spwa} for efficient DNN computing. As far as we know, there is still a lack of investigation of winograd convolution from the perspective of fault tolerance. In this work, we aim to investigate the fault tolerance of winograd-based convolution neural networks and take advantage of the improved neural network fault tolerance for soft error mitigation. 

\subsection{Reliability Analysis of Neural Networks}

Soft errors that are almost inevitable in existing large-scale VLSI designs typically manifest as bit flips and can propagate along with the neural network data flow. They may cause incorrect computing results and induce considerable accuracy loss. Quantifying the influence of soft errors and understanding the reliability of neural networks is an essential step to protect against the soft errors, especially for safety-critical applications like autonomous driving and robotics. There have been many prior works that evaluated the reliability of DNNs subjected to soft errors from different angles. For example, \cite{li2017understanding} studied the resilience of CNNs with different data types, values, data reuses, and types of layers to guide the fault-tolerant design. \cite{reagen2018ares} explored the relationship between fault rate and model accuracy under various setups such as quantization, layer types, and network structures. 
\cite{spyrou2022reliability} analyzed the impact of faults on SNNs with different parameters including splitters, routers, and neurons. Some of the reliability evaluations explored the reliability difference among components of neural networks such that they can be utilized to perform selective protection with less protection overhead \cite{mahmoud2021optimizing, ping2020sern, mahmoud2020hardnn}. More neural network reliability evaluation works can be found in recent surveys \cite{mittal2020survey} \cite{liu2022special}. 

Fault injection platforms or fault simulation are usually required for the reliability study of neural network processing. Neuron-level fault injection platforms such as TensorFI \cite{chen2020tensorfi}, PyTorchFI \cite{mahmoud2020pytorchfi}, Ares \cite{reagen2018ares} typically have bit errors injected to neurons or activations to model the influence of soft errors on neural network processing. They are widely utilized for the generality and much more convenient deployment. In contrast, some recent fault injection platforms have specific hardware architectures like GPUs \cite{tsai2021nvbitfi} \cite{hari2017sassifi} and neural network accelerators \cite{he2020fidelity} \cite{tan2023saca} considered for more accurate simulation. Compared to the neuron-level fault injection, they have higher fault simulation accuracy, but they are closely coupled with specific hardware architectures, which limits the generality of the analysis and slows down the simulation substantially.

\subsection{Fault Tolerance of Neural Networks}

To protect neural network processing against errors in silicon, many fault-tolerant approaches have been extensively studied.  Although conventional fault-tolerant design approaches such as triple modular redundancy, dual modular redundancy (DMR), ECC, and algorithm-based fault tolerance (ABFT) \cite{xu2021r2f, lee2022value, li2022efficient, ozen2019sanity} can potentially alleviate the influence of soft errors, they usually require considerable computing overhead. For instance, the authors in \cite{kao2022efficient} have an self-checking scheme integrated in each modular of a typical DMR approach such that each modular can detect errors by itself. In this case, the revised DMR can also recover when only one of the modular is faulty. Although the redundancy overhead is reduced substantially compared to a standard TMR, it still takes more than 100\% computing overhead and remains prohibitively expensive. In contrast, it is more flexible and cost-effective to take advantage of the inherent fault tolerance of the neural networks and alleviate the fault-tolerant design overhead. The authors in \cite{dutta2019codenet, ghavami2022fitact, hacene2019training, xu2019resilient} explored the fault tolerance of neural networks with retraining. Basically, it has the underlying hardware fault information learned and has the retrained model to adapt to the fault information. However, retraining is required for different fault scenarios and will induce considerable training cost. Several regularization techniques \cite{dey2017regularizing} \cite{ozen2021snr} have been proposed to constrain the range of weights and computing results in presence of hardware faults eventually. Unlike the works that retain the model architectures, the authors in \cite{li2020ftt} \cite{ning2021ftt} proposed to take fault tolerance into consideration in neural architecture search framework such that the searched network architecture fulfills both the fault tolerance metric and accuracy metric. More fault-tolerant approaches can also be found in recent surveys \cite{mittal2020survey} \cite{liu2022special} \cite{dave2022special} \cite{robust2020Mutlu}.

Existing works either require additional computing overhead or compromise the generality of the model. In this paper, we discover and evaluate the inherent fault tolerance of winograd convolution. With the fault tolerance of winograd convolution, we can achieve more cost-effective protection of neural network processing against soft errors without additional computing overhead nor accuracy penalty.

\begin{figure}[!t]
\vspace{-2em}
\centering
\includegraphics[width=3.4in]{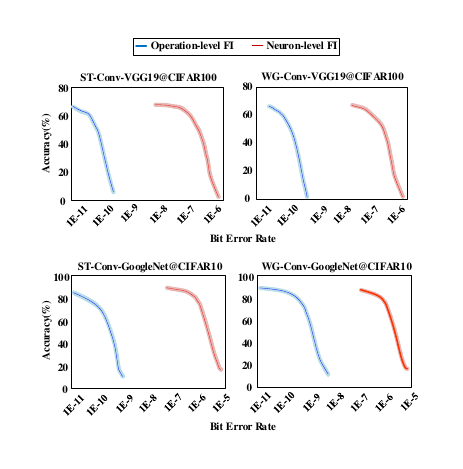}
\vspace{-2em} 
\caption{Model accuracy drop curves under different bit error rate (BER) using both the neuron-level fault injection (FI) and the operation-level FI. The accuracy drop curves obtained from different FI platforms are generally consistent while the shift between the curves are mainly caused by the different BER definitions.}
\vspace{-1em} 
\label{neural-ope-wise-fi}
\end{figure}

\section{Fault Injection Platform}
\label{sec:FI platform}

To address the reliability analysis problem of winograd convolution, we propose an operation-level fault injection platform for more fine-grained reliability analysis without compromising the generality of the fault injection nor slowing down the fault simulation speed too much. Specifically, it has random bit-flip errors injected to fine-grained primitive operations such as addition and multiplication rather than coarse-grained operations of entire neurons. At the same time, it retains the advantage of prior neuron-level fault injection and also targets general neural network processing rather than a specific computing engine. We implemented it on top of PyTorch and open sourced on github.

To illustrate the accuracy of the proposed operation-level fault injection, we take VGG19 on CIFAR-100 and GoogleNet on CIFAR-10 quantized with 16bit fixed point as an example, and compare the resulting model accuracy over that based on neuron-level fault injection. Specifically, we utilize bit error rate (BER) as the error metric. Although BER is originated from hardware error metric which refers to the bit flip error probability of a single memory bit, it is adapted to fit the soft error metric of neural network processing. Essentially, from a statistical perspective, BER represents the ratio of the total number of bit errors over the total number of neuron bits or operation bits in an neural network. For the neuron-level fault injection, BER is calculated based on neurons, following previous work\cite{reagen2018ares} \cite{mahmoud2020pytorchfi}, we inject random soft errors into the output neurons of each convolution layer to simulate the bit-flip error. For the operation-level fault injection, BER is calculated based on operations. The model accuracy of standard convolution and winograd convolution under different BER setups obtained using different fault injection platforms are presented in Figure \ref{neural-ope-wise-fi}. We present the results with error band for more comprehensive comparison in which the confidence interval is set to be 95\%. It can be seen that the trend of the model accuracy drop under different BER setups are generally similar. The difference of the two accuracy curves are mainly caused by the different BER setups. For example, for VGG19 and GoogleNet based on winograd convolution, the BER based on neuron-level fault injection are roughly $1.4\times10^3$ and $1.6\times10^3$ higher than the operation-level fault injection respectively. The number of primitive operations of VGG19 and GoogleNet are roughly $1.2\times10^3$ and $1.3\times10^3$ more than the number of neurons, respectively. It can be seen that the BER difference is roughly consistent with the difference of operation number and neuron number in both VGG19 and GoogleNet, which confirms the inherent consistence of the two fault injection methods.

\begin{figure}[!t]
\vspace{-2em}
\centering
\includegraphics[width=3.5in]{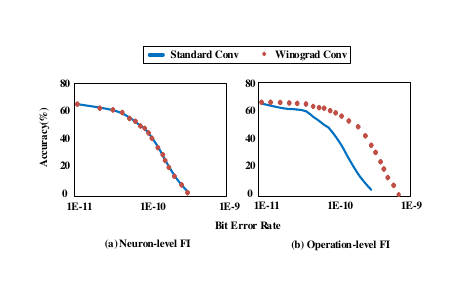}
\vspace{-3em} 
\caption{Neuron-level fault injection (FI) fails to differentiate the influence of soft errors on standard convolution and winograd convolution, while the operation-level FI with more fine-grained fault injection clearly shows the different model accuracy drop of standard convolution and winograd convolution under various error rate setups.}
\vspace{-1em} 
\label{diff_analy_meth}
\end{figure}

To enable more clear comparison, we have BER in neuron-level fault injection scaled to that defined in operation-level fault injection. In this case, we evaluate the accuracy of standard convolution and winograd convolution under different BER setups. The experiment result is shown in Figure \ref{diff_analy_meth}. It can be seen that there is no accuracy difference between standard convolution and winograd convolution under different bit error rate when using the neuron-level fault injection platform. It shows that the influence of soft error on standard convolution and winograd convolution cannot be distinguished as expected, because neuron-level fault injection is performed on neurons rather the neuron computing procedures. Nevertheless, the difference is clearly observed when the fine-grained operation-level fault injection is adopted. At the same time, we can observe that despite the difference in fault simulation accuracy, the general trend of the model accuracy under different bit error rate obtained from the two different fault injection platforms remains consistent.

\begin{figure}[!t]
\vspace{-2em}
\centering
\includegraphics[width=3.5in]{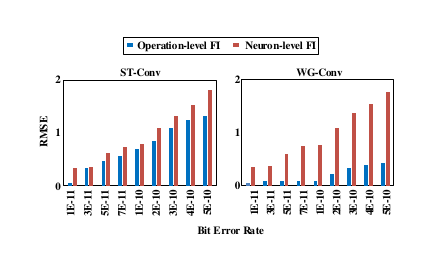}
\vspace{-3.5em} 
\caption{Comparison of the RMSE of standard convolution and winograd convolution with operation-level fault injection and neuron-level fault injection at the 11th layer of VGG19.}
\vspace{-1em} 
\label{rmse}
\end{figure}

On top of the model-wise comparison, we also compare the computing difference of a single neural network layer using operation-level fault injection and neuron-level fault injection respectively in presence of different bit flip errors. Specifically, we utilize root mean square error (RMSE) as the output variation metric and we take the 11th layer of VGG19 on CIFAR-100 as an example. The comparison is presented in Figure \ref{rmse}. It reveals that the RMSE obtained with operation-level fault injection and neuron-level fault injection of a standard convolution under different error rate is quite close to each other in general. In contrast, the RMSE differs considerably for winograd convolution and the RMSE obtained with operation-level fault injection is much lower. The main reason is that neuron-level fault injection fails to take advantage of the fault tolerance brought by winograd convolution.

\begin{figure}[!t]
\vspace{-2em}
\centering
\includegraphics[width=3.5in]{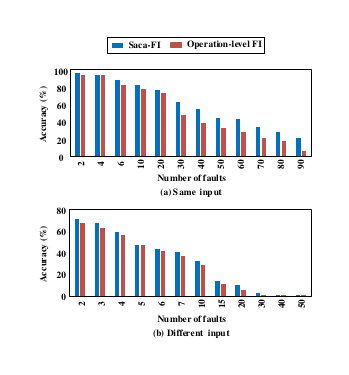}
\vspace{-3.5em}
\caption{Comparison between operation-level fault injection and Saca-FI that has architecture details considered in fault injection \cite{tan2023saca}.}
\vspace{-1em} 
\label{fig:compare_fi}
\end{figure}

In addition, we also have the proposed fault injection platform compared to fault injection platforms with hardware architecture details. While existing fault injection platforms with architecture details \cite{tsai2021nvbitfi} \cite{tan2023saca}  focus on the analysis of different consequences of errors and they are usually difficult to be applied for model accuracy evaluation directly, we revised a state-of-the-art fault injection platform based on scale-sim simulator \cite{tan2023saca} for the model accuracy analysis. We reuse the same floating point model VGG16 evaluated in \cite{tan2023saca} for the accuracy evaluation on ImageNet dataset. As errors in control logic are not considered in operation-level fault injection, we simply avoid the errors that will crash the control logic for a fair comparison. The model accuracy comparison under different BER setups is presented in Figure \ref{fig:compare_fi}. The accuracy measured in Figure \ref{fig:compare_fi}(a) refers to the percentage of correct inference over the total number of inference using the same input, while the accuracy used in Figure \ref{fig:compare_fi}(b) refers to the percentage of correct inference over the total number of inference using different inputs. It can be observed that the proposed operation-level fault injection shows quite similar accuracy drop with that evaluated with the fault injection in \cite{tan2023saca} in both cases, which confirms the fault simulation accuracy of operation-level fault injection. 

\begin{figure}[!t]
\vspace{-1em}
\centering
\includegraphics[width=3.5in]{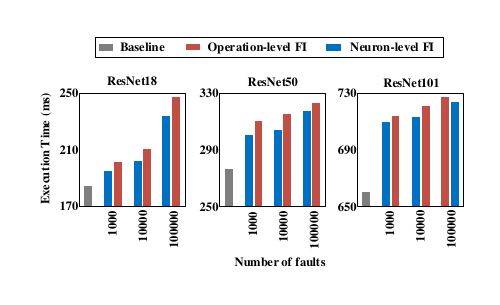}
\vspace{-3em} 
\caption{Fault simulation time of various neural network models in presence of different fault injection.}
\vspace{-1em} 
\label{diff_fi_time}
\end{figure}

Fault injection usually poses negative influence on neural network processing, we investigate the runtime of the proposed operation-level fault injection platform subjected to different errors on a typical PH402 SKU 200 GPU, and compare the runtime with that of a baseline framework without error injection and the neuron-level fault injection framework. In addition, we have a set of different ResNet models on ImageNet including ResNet18, ResNet50, ResNet101 utilized as the benchmark. The experiment result is presented in Figure \ref{diff_fi_time}. It can be observed that the operation-level fault injection is 1.07$\times$ and 1.03$\times$ slower than the baseline and neuron-level fault injection respectively. Particularly, this slowdown remains consistent across different model sizes, validating the scalability of the proposed fault injection platform over varying models.


\section{Fault Tolerance of Winograd DNNs}
\label{sec:evaluation}

Based on the proposed operation-level fault injection platform, we evaluate the fault tolerance of winograd-based DNNs from different granularities. First of all, we compare the fault tolerance of standard convolution and winograd convolution under different bit error rate. Then, we study standard convolution and winograd convolution in terms of layer granularity. Finally, we investigate the impact of soft errors on different types of operations in both standard convolution and winograd convolution.

\subsection{Evaluation Setup}

The major experiment setups include datasets and models benchmark, fault models, fault injection framework, and hardware platforms.

\begin{figure*}[!t]
\vspace{-1em}
\centering
\includegraphics[width=7in]{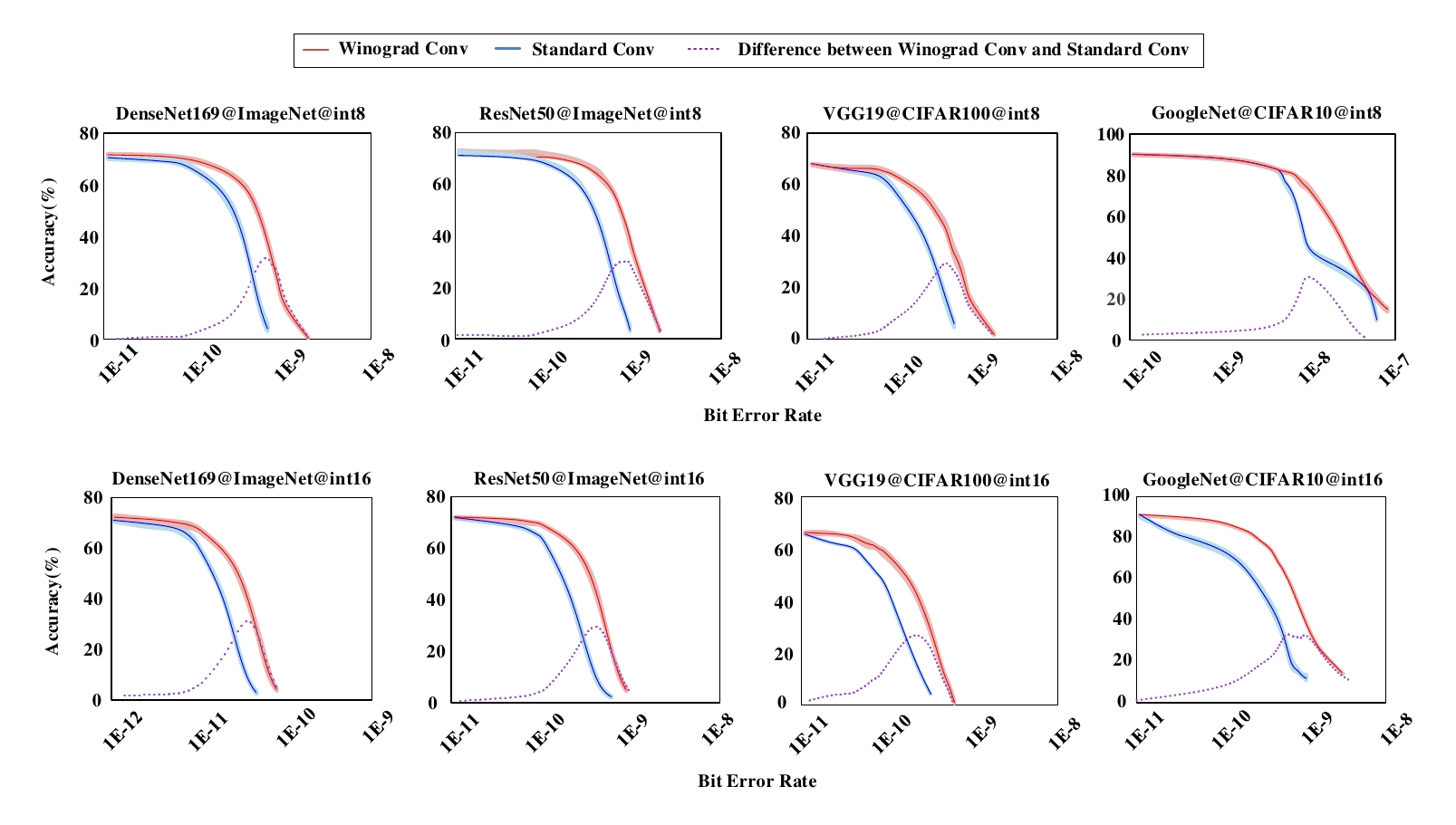}
\vspace{-1em}
\caption{Accuracy of benchmark neural networks calculated with standard convolution and winograd convolution.}
\vspace{-1em} 
\label{diff_net_conv_win_8_16_a}
\end{figure*}

\textbf{Datasets And Models.} In this evaluation, we  use DenseNet169 on ImageNet, ResNet50 on ImageNet, VGG19 on CIFAR-100, and GoogleNet on CIFAR-10 as the benchmark neural networks. Each neural network is quantized to a 8bit fixed point version and a 16bit fixed point version respectively. 

\textbf{Fault Models.} We utilize bit flip to characterize typical soft errors. Particularly, following prior reliability analysis work \cite{reagen2018ares}\cite{li2017understanding} , we also use bit error rate (BER), which represents the ratio of bit flip errors over the total number of bits of operations in the model as the soft error intensity metric.

\textbf{Error Injection.} We adopt the operation-level fault injection platform proposed in Section \ref{sec:FI platform} for the reliability evaluation. It only covers the convolution layer in the model.

\textbf{Hardware Platforms.} All the evaluation experiments are performed on a server equipped with two 24-core@2.5GHz Intel Xeon processors, 512GB memory, and four PH402 SKU 200 GPU cards.

\subsection{Network-wise Fault Tolerance Evaluation}
\label{sec:network}

In this sub section, we evaluated the accuracy of the benchmark neural networks implemented with standard convolution and winograd convolution under different bit error rate setups ranging from 0 to 1E-7, and utilize the accuracy degradation curves to characterize the resilience of the models subjected to bit flip errors. 

The experiment result is shown in Figure \ref{diff_net_conv_win_8_16_a}. It can be observed that the general trend of the accuracy curves of different models are similar. Basically, when the bit error rate is low, most of the computing errors can be tolerated and the model accuracy generally remains steady or drops slightly. When the bit error rate further increases, the model accuracy drops sharply as the distributed errors reach certain threshold and corrupt the models. In addition, we notice that different model architectures generally exhibit different accuracy degradation curves. We also observe that under each model architecture, neural network implemented with winograd convolution generally shows significant higher accuracy compared to that implemented with standard convolution. As shown in the dotted line in the figure, winograd convolution can improve the accuracy by up to 40\% compared with standard convolution. The main reason is that winograd convolution \cite{lavin2016fast} can reduce the arithmetic complexity by 2.25$\times$ compared to standard convolution, and the total number of bit flip errors of winograd convolution within an inference will be much less than that of standard convolution due to the shorter runtime. Hence, the error induced computing variation will be less significant. While the performance advantage of winograd convolution over standard convolution depends on the neural network sizes, the total number of errors injected also varies for different neural network models and leads to different accuracy loss accordingly. In addition, we find that models quantized with int16 are more vulnerable than that with int8, because bit flip for int16 can cause larger data variation on average.

\subsection{Layer-wise Fault Tolerance Evaluation} \label{sec:layer-vulnerability}
\begin{figure}[!t]
\vspace{-1em}
\centering
\includegraphics[width=3.3in]{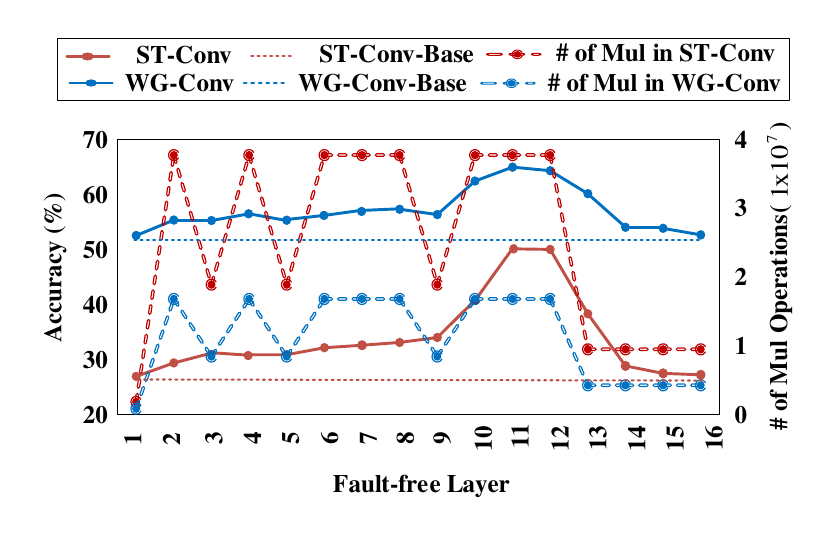}
\vspace{-1em}
\caption{Accuracy of VGG19 on CIFAR-100 with one fault-free layer while the rest of layers are injected using operation-level fault injection. We have the neural network implemented with standard convolution and winograd convolution respectively. The base accuracy refers to the occasion when all the neural network layers are injected with the same bit error rate.}
\vspace{-1em} 
\label{diff_layer_1}
\end{figure}

To gain insight of the fault tolerance of winograd DNNs, we conduct layer-wise fault analysis of the neural networks with both winograd convolution and standard convolution, which is also an important basis for selective model protection. To characterize a layer fault tolerance of neural networks, as shown in Figure \ref{diff_layer_1}, we take VGG19 on CIFAR-100 as an example to implement layer-wise fault tolerance evaluation. we choose fault injection with a baseline accuracy loss 15\% for winograd convolution. We use the model accuracy of standard convolution and winograd convolution at certain bit error rate as the baseline, denoted as ST-Conv-Base and WG-Conv-Base, respectively. We inject random bit errors into the operation of the entire neural network except the under evaluated layer to simulate model accuracy after protecting the under evaluated layer to obtain layer-wise model accuracy of the entire neural network using standard convolution and winograd convolution, denoted as ST-Conv and WG-Conv. The model accuracy difference of that obtained model accuracy after protecting the layer under evaluation relative to the baseline represents the fault sensitivity of the layer under evaluation. In general, larger accuracy difference indicates that higher model accuracy can be recovered from the baseline. Hence, the corresponding layer is more critical to the fault tolerance of the entire neural network. In addition, the variables \# of Mul in ST-Conv and \# of Mul in WG-Conv denote the number of multiplication operations involved in each layer of the standard convolution and winograd convolution models, respectively. These variables are employed to demonstrate the correlation between the fault sensitivity of the evaluated layer and the number of multiplication operations under operation-level fault injection.

The experiment result is shown in Figure \ref{diff_layer_1}. It reveals the sensitivity of the different layers of the neural network. As can be seen from the figure, there is a large difference in sensitivity between the layers of the standard convolution and winograd convolution neural networks. Among them, the centering layers of both standard convolution and winograd convolution neural network are more sensitive to the bit errors, while the layers in the beginning and the end of the network are less sensitive. The accuracy difference between layers of winograd convolution is up to 12.34\%, and the accuracy difference between layers of standard convolution is up to 27.21\%. It is probably because of the difference in operation number involved in each layer, as the layer-wise model accuracy is roughly consistent with the number of multiplication operations involved in each layer according to Figure \ref{diff_layer_1}. Basically, the layers with more operations are likely to induce higher accuracy improvement. In addition, the layer-wise accuracy of the neural network implemented with winograd convolution is generally much higher than that implemented with standard convolution. The trend of the layer-wise accuracy is roughly consistent for neural networks implemented with both standard convolution and winograd convolution.

\subsection{Operation Type Fault Tolerance Evaluation} \label{sec:type-vulnerability}
As winograd greatly changes the number of multiplication and addition in neural networks, we also investigate the fault tolerance from the perspective of operation types. Evaluation metrics are similar to layer-wise fault tolerance evaluation metrics. Assume that the entire model is exposed to random bit errors, then the model accuracy at which the multiplication operations are kept fault-free can be used to measure the sensitivity of the multiplication operations to bit errors. Higher accuracy indicates that these operations are more vulnerable and need to be protected with higher priority. Similarly, the sensitivity of operations in different neural networks under different bit error rate can be obtained. 

\begin{figure*}[!t]
\vspace{-1em}
\centering
\includegraphics[width=6.5in]{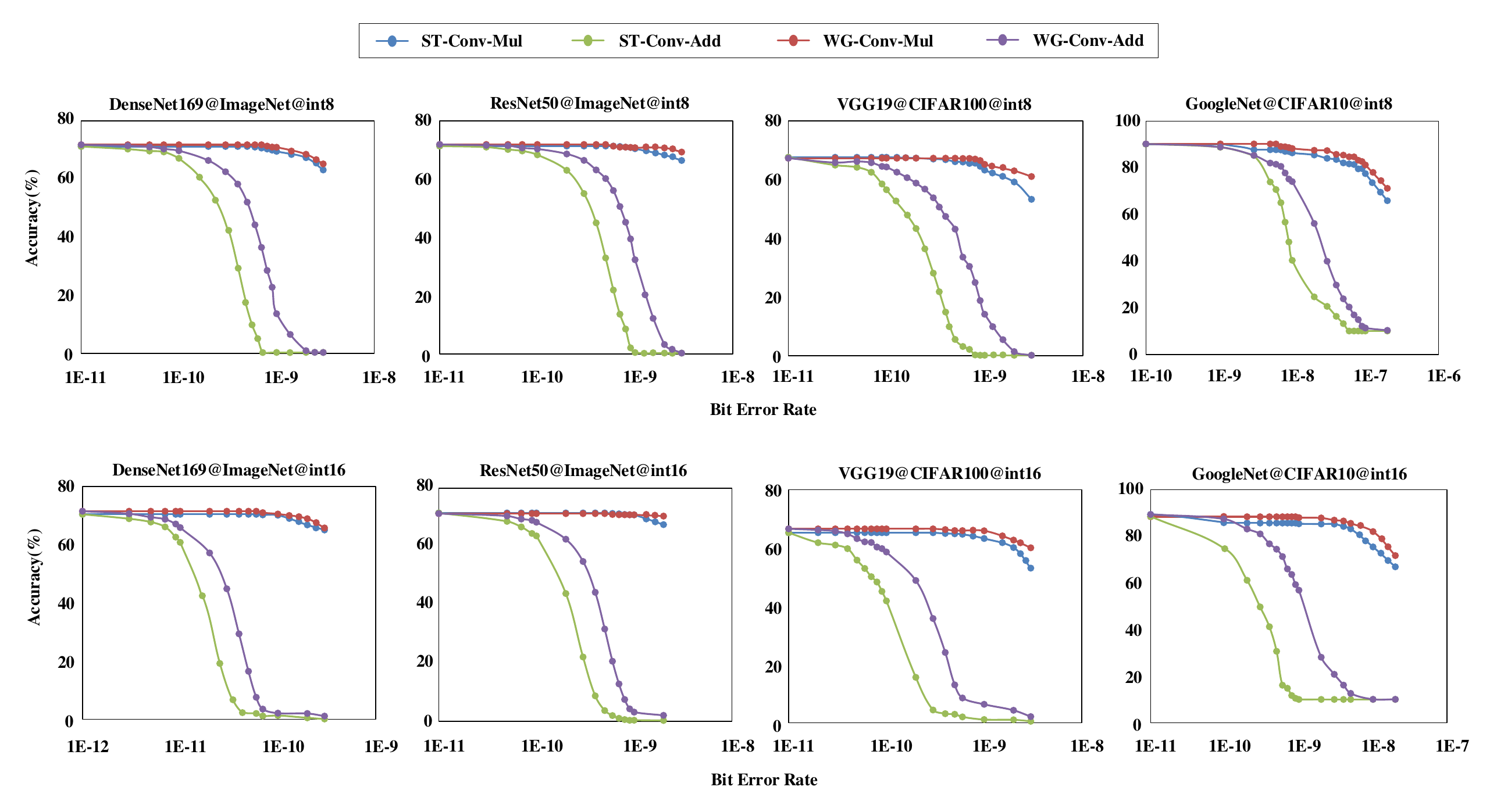}
\vspace{-1em}

\caption{Accuracy of standard and winograd neural networks with fault-free addition or fault-free multiplication under different bit error rate.}

\label{diff_ope}
\end{figure*}

The experiment result is shown in Figure \ref{diff_ope}. Note that ST-Conv-Add and ST-Conv-Mul represent the accuracy of fault-free  addition and fault-free multiplication in standard convolution respectively, while WG-Conv-Add and WG-Conv-Mul represent the accuracy of fault-free addition and fault-free multiplication in winograd convolution respectively. The experiment result shows that multiplication operations are more vulnerable than addition operations in standard convolution and winograd convolution of different neural networks under various bit error rate. In addition, we notice that the accuracy of WG-Conv-Add with more addition included is generally higher than that of ST-Conv-Add, but addition remains much less important to the accuracy of the entire neural network. In contrast, WG-Conv-Mul achieves comparable accuracy to ST-Conv-Mul at most bit error rates, but since winograd convolution have much fewer multiplication operations than standard convolution. This shows that compared to standard convolution, winograd convolution only needs to protect fewer multiplication operations to achieve the same accuracy, which makes winograd convolution more cost-effective for protection than standard convolution.

\section{Exploring Winograd for Cost-effective Fault Tolerance}
We already demonstrate the significant fault tolerance advantage of winograd convolution in Section \ref{sec:evaluation}. However, winograd convolution can still suffer dramatic accuracy loss when soft errors exceed the fault-tolerance capability. In this case, we still need more intensive protection, and we mainly investigate how winograd convolution can be explored for more cost-effective fault tolerance using existing fault-tolerant design approaches in this section. Specifically, we take TMR and fault-tolerant DNNs based on fault-aware retraining and constrained activation functions as typical examples, and explore the use of winograd convolution for less fault-tolerant design overhead.

\subsection{Winograd Convolution and TMR}

TMR is a classical approach to mitigate soft errors in silicon. Prior layer-wise TMR \cite{xu2021r2f} for DNNs protects only a fraction of layers based on the vulnerability of the layers. Although more fine-grained protection is also possible, it remains difficult to separate the different operation types because the different operation types are closely coupled especially for DNN accelerators. We notice that winograd convolution splits the multiplication and addition operations considerably, which enables TMR protection across different operation types and layers without altering the computing patterns. On top of winograd convolution, we propose a fine-grained TMR protection approach as shown in Algorithm 1, which priorities the protection of a set of continuous operations across both the layers and operation types. First of all, we divide the operations into continuous segments with the same number of operations. Suppose the protection granularity of the basic processing segment is $m$, and the total number of operations of the model is $M$, we get $\lceil M / m \rceil$ basic segments to be protected (line 2). Then, we select the segments for protection iteratively. Specifically, we utilize the vulnerability factor of the basic computing segments $S_i$ as the protection priority metric. The vulnerability factor $V_i$ is defined as the model accuracy improvement between models with the target computing segment protected and models without any protection (lines 3-8). Computing segments with higher vulnerability factor are preferred for the protection because they promise higher model accuracy improvement. As different computing segments may be correlated, we gradually add the segments for protection and investigate the accuracy at the end of each iteration. The procedure will stop when the model accuracy reaches the design goal of accuracy (i.e., $ACC$) such that the TMR protection overhead can be constrained(line 10-14). To simplify the protection granularity characterization, we utilize the ratio $P=nm/M$ instead. Larger $P$ indicates more coarse-grained protection granularity and $P$  can be used to compromise between the TMR overhead and model accuracy. 

\algdef{SE}[DOWHILE]{Do}{doWhile}{\algorithmicdo}[1]{\algorithmicwhile\ #1}%
\begin{algorithm}[!ht]
\caption{Fine-grained TMR Protection Algorithm}
\hspace*{0.02in} {\bf Input:} The total number of neural network computing operations $M$, basic processing segment size $m$, the design target accuracy of model $ACC$.\\
\hspace*{0.02in} {\bf Output:} The total number of protected segments $n$.
\begin{algorithmic}[1]
\State Initialize $n$ ← $0$
\State Number of segments $N$ ← $\lceil M / m \rceil$ 
\For {$i$ = 1 to $N$}
      \State $S_i$ is the $i$th computing segment. 
      \State $acc_{prot}$ ← $MeasureProtectedAcc(\{S_i\})$
      \State $acc_{raw}$ ← $MeasureProtectedAcc(\varnothing)$
      \State Let vulnerability $V_i$ ← $acc_{prot} - acc_{raw}$
\EndFor
\State Sort $S_i$ in descending order by $V_i$
\Do
   \State $P_n$ ← $\{S_1, S_2, \cdots, S_{n}\}$
   \State $acc$ ← $MeasureProtectedAcc(P_n)$
   \State $n$ ← $n+1$
\doWhile{$acc < ACC$}
\State \Return $n$

\end{algorithmic}  
\end{algorithm}

We have the proposed TMR protection approach applied to three different neural network implementations including standard convolution (ST-Conv), winograd convolution without being aware of the fault tolerance (WG-Conv-W/O-AFT), and winograd convolution with being aware of the fault tolerance (WG-Conv-W/AFT). Note that WG-Conv-W/O-AFT utilizes the same TMR protection option with ST-Conv because it is not aware of the fault tolerance of winograd convolution. The major difference between WG-Conv-W/O-AFT and ST-Conv is that WG-Conv-W/O-AFT conducts the neural network processing and protection on top of winograd convolution, while ST-Conv conducts on standard convolution. In contrast, WG-Conv-W/AFT conducts TMR protection on top of winograd convolution using fault-tolerant analysis results of winograd convolution. The experiments use VGG19 quantized with int16 on CIFAR-100 as the benchmark example and its original model accuracy is 72.6\%. The bit error rate is set to 30\% accuracy loss.

\begin{figure}[!t]
\vspace{-1em}
\centering
\includegraphics[width=3.45in]{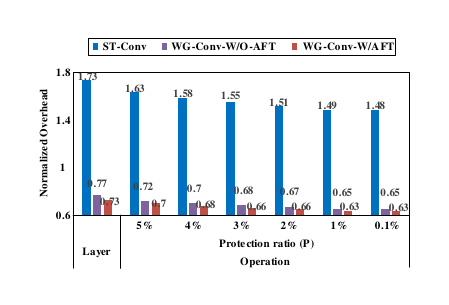}
\vspace{-3em}
\caption{Normalized TMR overhead of different approaches at different protection ratio $P$ when the accuracy target is 70\%.}
\vspace{-1em} 
\label{tmr_grain}
\end{figure}

We evaluate the TMR overhead of the three different convolution processing approaches including ST-Conv, WG-Conv-W/O-AFT, and WG-Conv-W/AFT at different protection ratio i.e. $P$ setups. $P$ ranges from 5\% to 0.1\% and the target model accuracy is set to be 70\%. All the TMR overhead is normalized to that of ST-Conv. We utilize the number of primitive operations such as addition and multiplication to measure the computing overhead. According to the comparison in \cite{horowitz20141}, multiplication is generally more expensive than addition, so we have the operations weighted for a fair computing overhead comparison. The overhead of an multiplication (int8) is set to be 6.67$\times$ higher over that of an add operation (int8). When the proposed fine-grained TMR protection is applied, the TMR overhead based on different convolution processing approaches is shown in Figure \ref{tmr_grain}. It can be observed that the TMR overhead generally gets lower given more fine-grained TMR protection granularity as expected. Compared to conventional layer-wise TMR protection, the proposed fine-grained TMR reduces the TMR overhead by 3.84\% to 15.3\% depending on the protection granularity. The benefits of fine-grained TMR protection gets saturated when $P$ is closer to 1\%. The is probably attributed to the limited resolution of the vulnerability factor calculation based on fault injection. When comparing the different convolution processing approaches, we notice that WG-Conv-W/AFT requires the least TMR overhead under all the different protection granularity setups. Specifically, WG-Conv-W/O-AFT shows 55.77\% less TMR overhead than ST-Conv on average, which is mainly attributed to the reduced computing overhead of fine-grained winograd convolution. When the fault tolerance capability of winograd convolution is considered, TMR computing overhead can be further reduced by 17.24\% without compromising the model accuracy.

\begin{figure}[tb]
\vspace{-1em}
\centering
\includegraphics[width=3.45in]{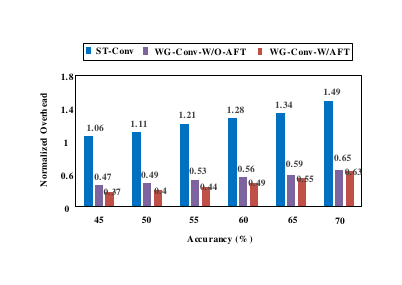}
\vspace{-3.5em}
\caption{Normalized TMR overhead under different accuracy when protection ratio is 1\%.}
\label{tmr}
\end{figure}

We also evaluate the TMR overhead under different design goals ranging from 45\% to 70\%. The protection granularity ratio $P$ is set to be 1\%. The TMR overhead is normalized to that of ST-Conv based on the number of primitive operations. The experiment result is presented in Figure \ref{tmr}. It can be observed that TMR overhead can be reduced dramatically when lower model accuracy is required and less DNN processing needs to be protected. Although the computing benefit is mainly attributed to the reduced multiplication operations in winograd convolution, the fault tolerance of winograd convolution also contributes substantially especially when the target model accuracy is less stringent. For instance, when the target model accuracy is 45\%, WG-Conv-W/AFT that fully explores the winograd convolution fault tolerance reduces the TMR overhead by 20\% compared to WG-Conv-W/O-AFT.

\subsection{Winograd convolution and fault-tolerant DNNs}

\begin{figure*}[!t]
\vspace{1em}
\centering
\includegraphics[width=6.8in]{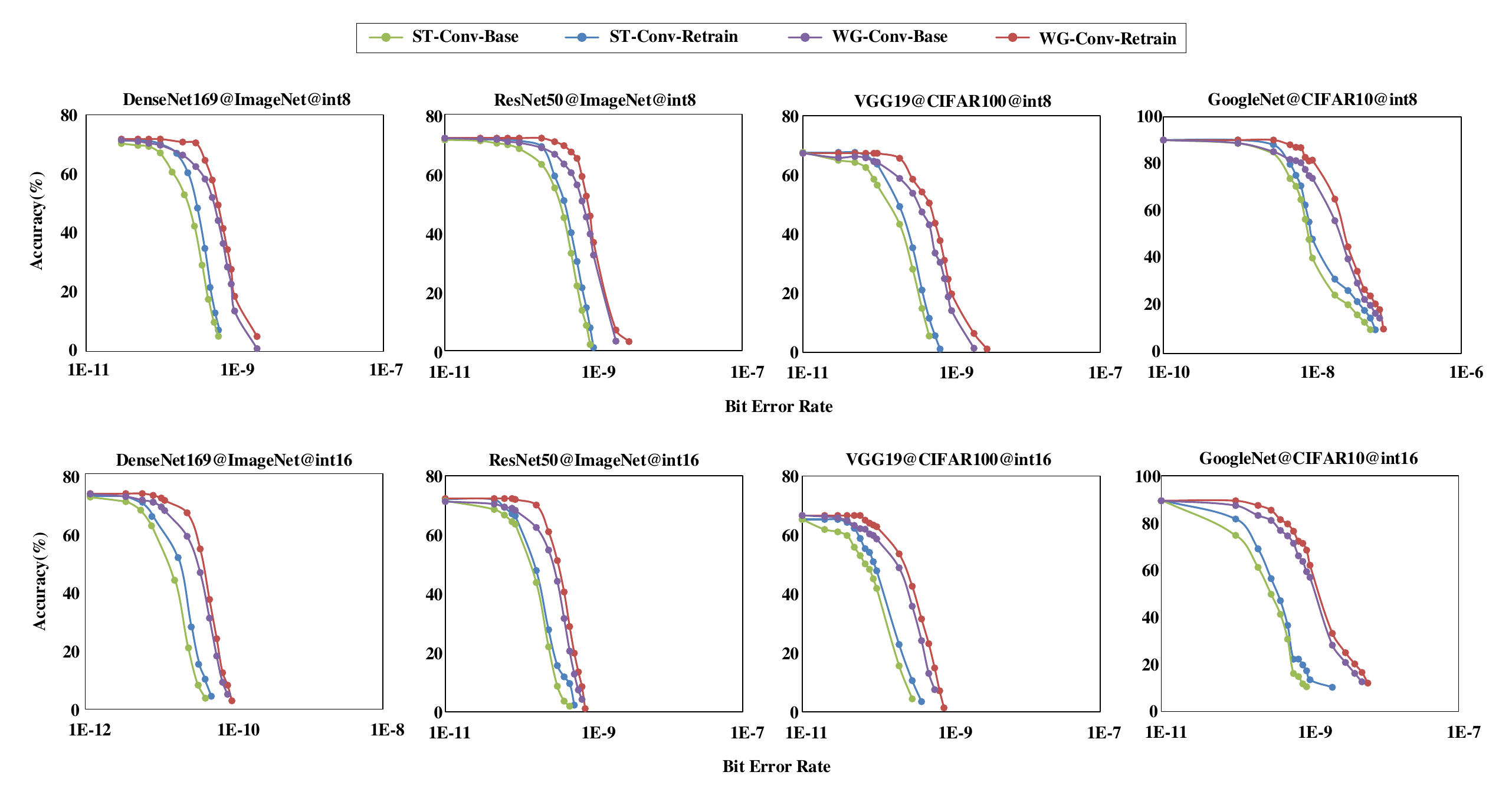}
\vspace{-1em}
\caption{Fault-aware retraining for standard convolution and winograd convolution on different networks under different bit error rates.}
\vspace{-1em} 
\label{retrain}
\end{figure*}

Different from classical TMR, there are also fault-tolerant design approaches including fault-aware retraining and constrained activation functions that investigate the inherent fault tolerance of DNNs. In this section, we mainly explore whether the fault-tolerant DNNs can benefit from the fault tolerance of winograd convolution. Specifically, fault-tolerant retraining generally takes soft errors induced computing variation into loss calculation such that the retrained model have both data and computing errors involved in the model. As for constrained activation functions, it mainly revises the activation function such as Relu with additional range constraints to filter out neurons with very large or small values which are probably caused by soft errors. The range is obtained through profiling of the maximum and minimum data of neurons in the DNNs. Essentially, it suppresses the influence of soft errors on DNN processing. 

\begin{figure*}[!t]

\centering
\includegraphics[width=6.8in]{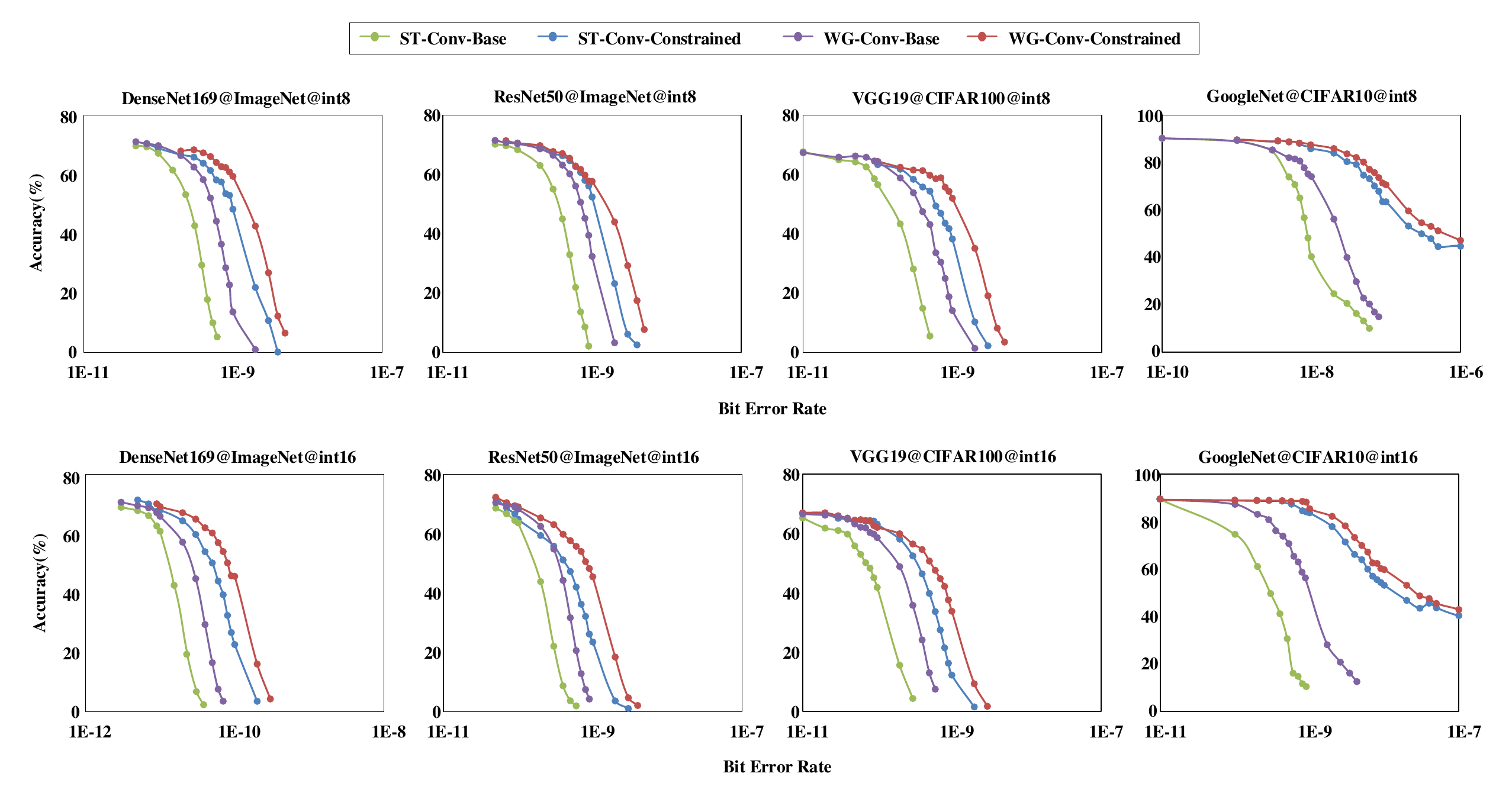}
\vspace{-1em}
\caption{Constrained activation function for standard convolution and winograd convolution on different networks under different bit error rates.}
\vspace{-1em} 
\label{range_based}
\end{figure*}

In this section, we apply winograd convolution to the fault-tolerant DNNs based on fault-aware training and constrained activation functions, and compare them with the fault-tolerant DNNs without using winograd conversion under various bit error rate setups. The comparison is shown in Figure \ref{retrain} and Figure \ref{range_based}. In general, winograd convolution can further enhance the model accuracy on top of existing fault-tolerant DNNs despite the bit error rate setups, DNN models, and quantization data width. Particularly, as shown in Figure \ref{retrain}, we notice that winograd based DNN processing generally shows comparable model accuracy improvement when the fault-aware training is applied under various setups, which roughly demonstrates orthogonal fault tolerance capability between winograd convolution and fault-aware training. As for the constrained activation functions in Figure \ref{range_based}, we observe that winograd convolution also clearly improves the fault tolerance of DNNs under all the different setups. Nevertheless, the improvement is generally higher for the baseline models while the improvement is limited for the DNNs with constrained activation functions. The main reason is probably because the constrained activation functions already enhance the model significantly and leaves limited space for winograd convolution, but winograd processing is still great beneficial and requires less computing at the same time. In summary, we can conclude that winograd based DNN processing provides a unique angle of fault-tolerant design and can be integrated with the major fault-tolerant design approaches for DNNs.

\section{Conclusion}

This paper discovers and studies the fault tolerance of winograd-based DNNs comprehensively, and evaluates its fault tolerance from different granularities such as models, layers and operation types for the first time. With the evaluation, we further explore the use of winograd fault tolerance for cost-effective fault-tolerant designs. We propose a fine-grained TMR-based redundancy approach for DNNs that can span not only different layers but also different operation types. According to our experiments, the TMR computing overhead of winograd convolution is significantly reduced without compromising the accuracy. Additionally, we incorporate winograd convolution with classical fault-tolerant neural network approaches including fault-aware retraining and constrained activation functions, and verify its orthogonality with other classical fault-tolerant approaches.

\bibliographystyle{ieeetr}
\bibliography{ref}

\vfill
\end{document}